\documentclass[times,twocolumn,final,authoryear]{elsarticle}
\usepackage[caption=false,font=normalsize,labelfont=sf,textfont=sf]{subfig}
\usepackage{ycviu}
\usepackage{framed,multirow}
\usepackage{hyperref}
\usepackage{booktabs}
\usepackage{multirow}

\usepackage{amssymb}
\usepackage{latexsym}

\usepackage{url}
\usepackage{xcolor}
\definecolor{newcolor}{rgb}{.8,.349,.1}

\journal{Computer Vision and Image Understanding}

\begin{document}

\thispagestyle{empty}
                                                             

\begin{frontmatter}

\title{Multi-cue adaptive emotion recognition network}

\author[1]{Willams \snm{Costa}\corref{cor1}} 
\cortext[cor1]{Corresponding author:}
\ead{wlc2@cin.ufpe.br}
\author[2,3]{David \snm{Mac\^edo}}
\author[2,4]{Cleber \snm{Zanchettin}}
\author[1,5]{Lucas \snm{Figueiredo}}
\author[1]{Veronicca \snm{Teichrieb}}

\address[1]{Voxar Labs, Centro de Inform\'atica, Universidade Federal de Pernambuco, Recife, Brasil}
\address[2]{Centro de Inform\'atica, Universidade Federal de Pernambuco, Recife, Brasil}
\address[3]{Montreal Institute for Learning Algorithms, University of Montreal, Quebec, Canada}
\address[4]{Department of Chemical and Biological Engineering, Northwestern University, Evanston, United States of America}
\address[5]{Unidade Acadêmica de Belo Jardim, Universidade Federal Rural de Pernambuco, Belo Jardim, Brasil}

\received{1 May 2013}
\finalform{10 May 2013}
\accepted{13 May 2013}
\availableonline{15 May 2013}
\communicated{S. Sarkar}

\begin{abstract}
Expressing and identifying emotions through facial and physical expressions is a significant part of social interaction. The computer task for identifying emotions and allowing a more natural interaction between humans and machines is emotion recognition. The common approaches for emotion recognition focus on analyzing facial expressions and requires the automatic localization of the face in the evaluated scene. Although these approaches can classify emotion in controlled scenarios, such techniques are limited when dealing with unconstrained daily interactions. We propose a new direction for emotion recognition based on adaptive multi-cues using face and information from context and body poses. We evaluate the proposed approach in the CAER-S dataset, considering different components in a pipeline that is 21.49\% better than the current state-of-the-art method.
\end{abstract}

\begin{keyword}
\MSC 41A05\sep 41A10\sep 65D05\sep 65D17
\KWD Keyword1\sep Keyword2\sep Keyword3

\end{keyword}

\end{frontmatter}


\section{Introduction}
\label{intro}
Communication plays a crucial role in our social experiences. By conveying our thoughts and feelings, we can create social links. The human communication comprises verbal communication, in which we express ourselves using speech, and nonverbal communication, which refers to how our body expresses our feelings using facial expressions, gaze, gestures, and body language. Surprisingly, the nonverbal communication constitutes up to half of what we are communicating \citep{patel2014body}. Moreover, humans can interpret it even in an unconscious way.

Recent approaches for improving human-computer interaction (HCI) are shifting the focus from computer-centered domain to user-centered applications \citep{rouast2019deep}. In this sense, understanding emotional responses from users can make artificial systems change the context of what they are communicating and trigger different actions \citep{valli2008}, enabling different HCI applications such as education \citep{ref1_edu}, entertainment \citep{ref1_enter}, virtual reality \citep{ref1_vr}, mental health monitoring \citep{ref1_mentalhealth} and general healthcare \citep{ref1_health}.

Current research on emotion recognition is focused on facial expressions, motivated by the many discriminative features present on the human face. However, research on behavioral psychology indicates that one modality of nonverbal communication is not necessarily superior to the other on perceiving feelings and intents \citep{andreaaffective}. 

Still, estimating the correct user's feelings using techniques based only on facial expressions is a challenging task. For example, such methods do not perform well on unconstrained, daily situations (e.g., social interactions), mainly due to the need for a specific alignment of the user's face with the camera capturing the action \citep{ck}. Some methods boost the recognition performance by leveraging context information, such as gesture and context background, to allow more unconstrained scenarios \citep{caer, kosti2017emotion, kosti2019context, randhavane2019identifying, randhavane2019liar}. However, investigating other nonverbal cues, such as body language and the combination of different approaches, is yet an open topic in HCI research.

Recent works extract features from gait estimation, such as arm swinging, long strides, foot landing force, and posture to identify sentiment on videos. However, they also do not work well in unconstrained situations, obliging the user to face directly the camera \citep{randhavane2019identifying, randhavane2019liar, bhattacharya2020step, crenn2016body}.

\begin{figure}[!h]
\centering
\subfloat[]{\includegraphics[width=0.25\columnwidth]{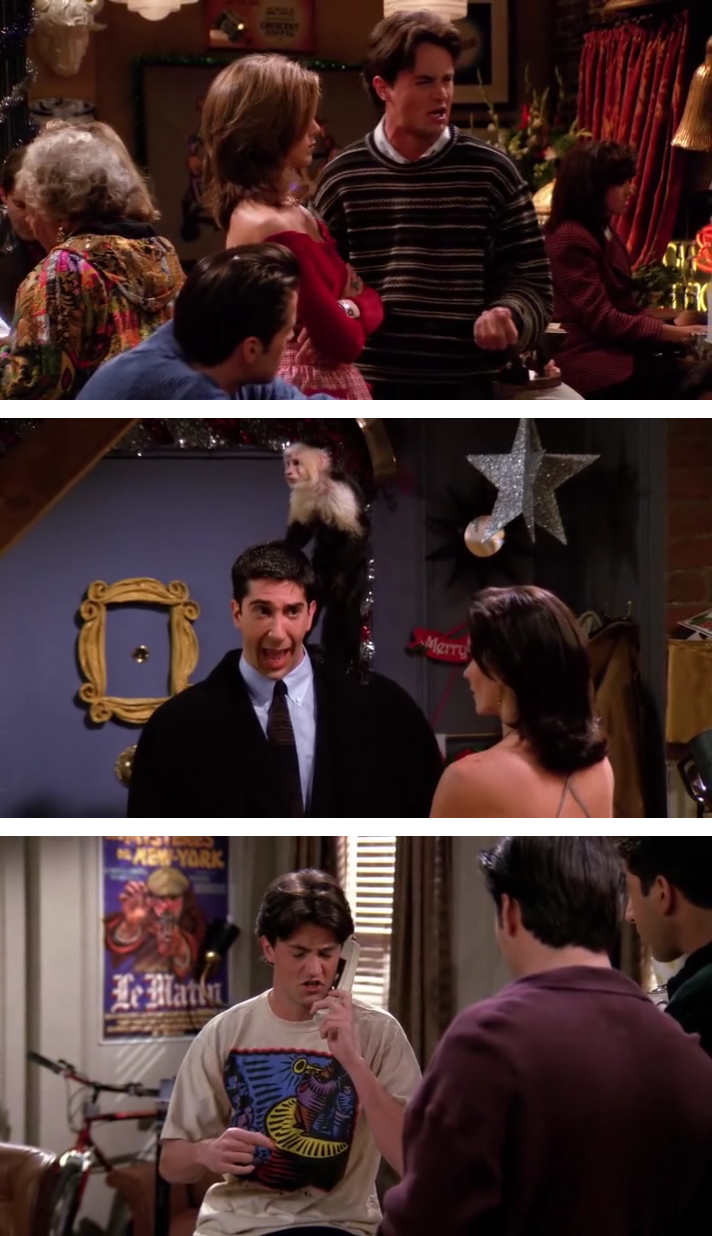}%
\label{fig_first_case}}
\hfil
\subfloat[]{\includegraphics[width=0.25\columnwidth]{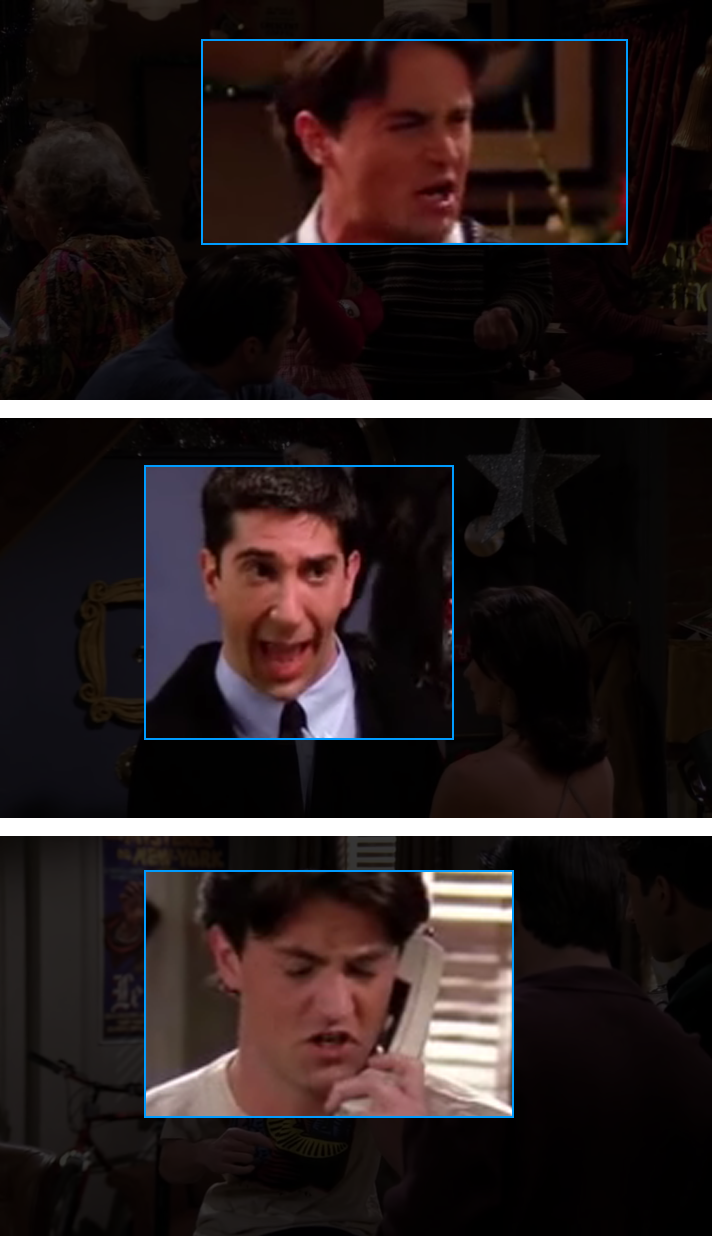}%
\label{fig_first_case}}
\hfil
\subfloat[]{\includegraphics[width=0.25\columnwidth]{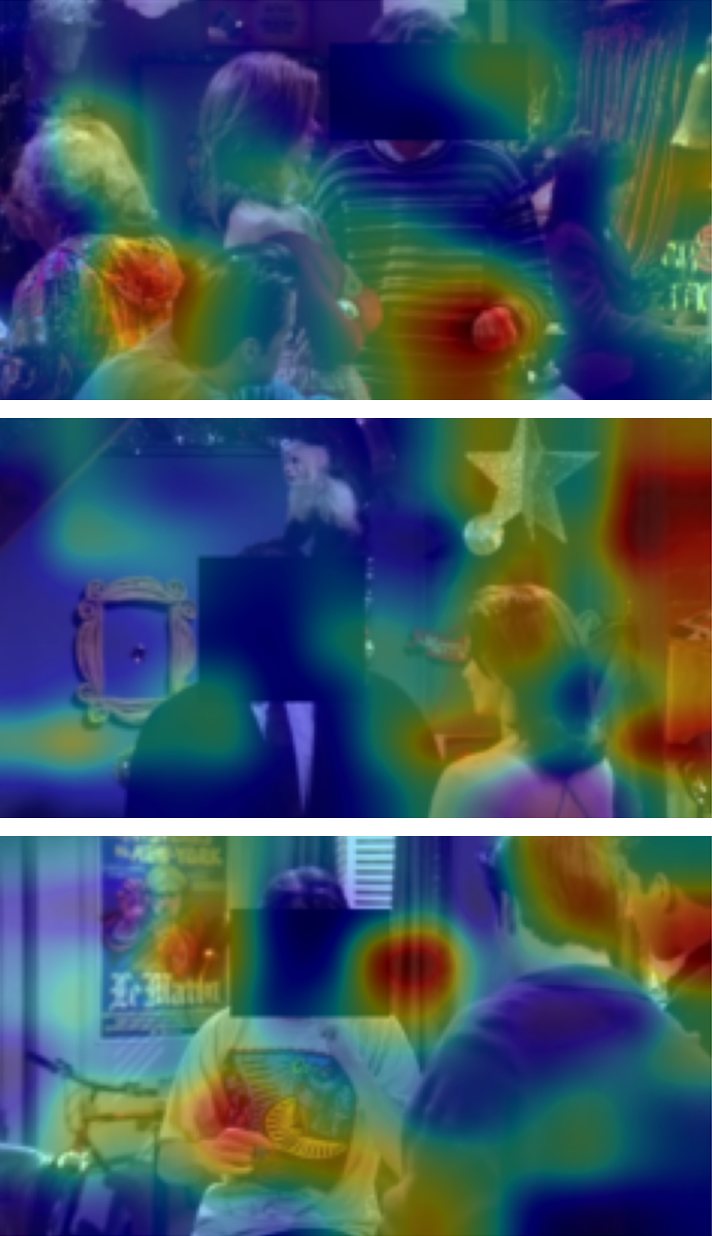}%
\label{fig_first_case}}
\hfil
\subfloat[]{\includegraphics[width=0.25\columnwidth]{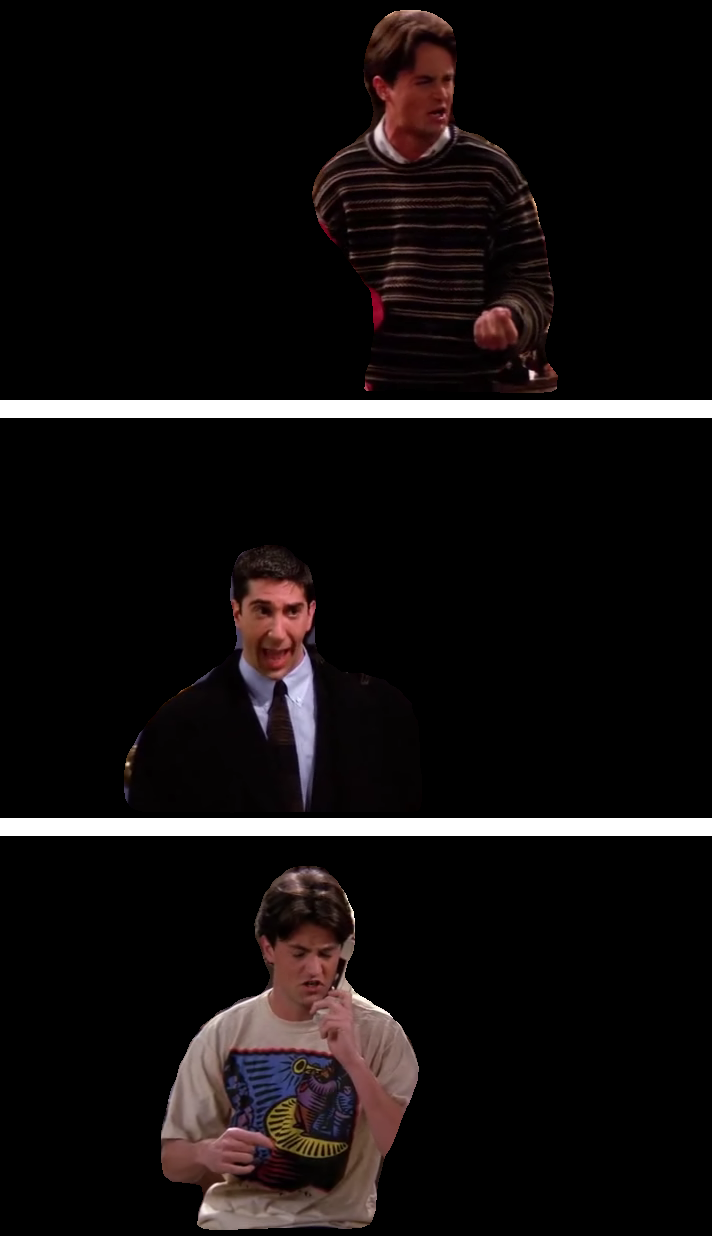}%
\label{fig_second_case}}
\caption{Overview of our solution. Given (a) images containing people, emotion recognition consists of understanding the emotion of a human being. Approaches for this task usually focus on the study of facial expressions, motivated by the idea that the most expressive non-verbal communication form is through facial expressions, as presented in (b). We propose a \textit{face encoding stream}, which can extract facial features and contribute as one of many input cues. Another relevant information is context since it can directly affect how one feels towards the environment. As we show in (c), we also propose a \textit{context encoding stream} powered by self-calibrated convolutions that can extract meaningful information from context, which is another input for our method. Finally, as we show in (d), another important form of non-verbal communication is body language and body pose, and we also consider it by proposing a \textit{body encoding stream} that extracts information regarding body pose.\\[-5mm]}
\label{fig_sim}
\end{figure}

In this work, we investigate the use of multiple cues that correspond to nonverbal communication, namely facial expressions, body language, and context, to recognize emotion in unconstrained scenarios. Fig.~\ref{fig_sim} presents an overview of the proposed solution, named Multi-Cue Adaptive Emotion Recognition Network (MCAER-Net).

\begin{figure*}[t!]
	\centering
	\includegraphics[width=\textwidth]{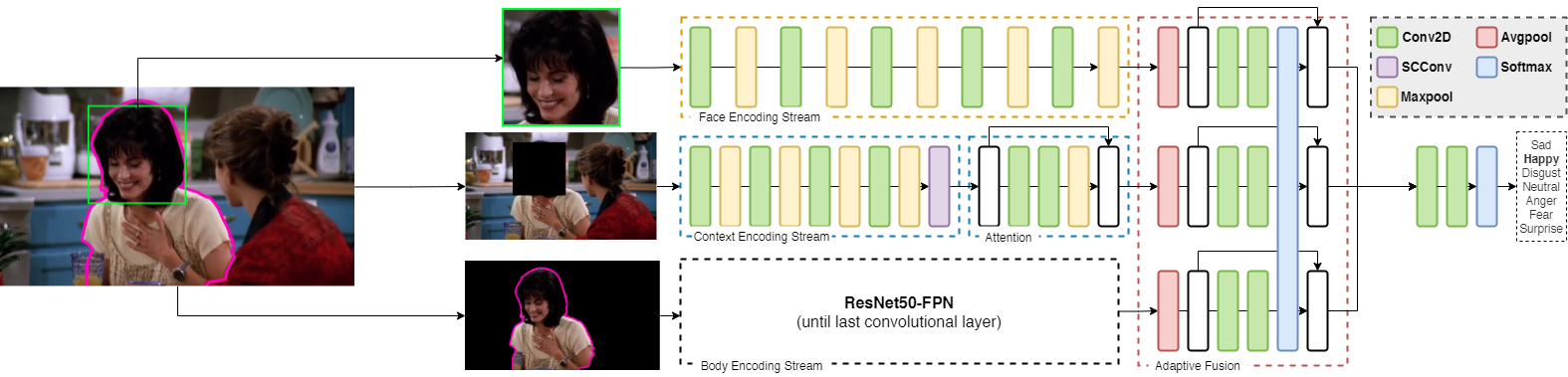}
	\caption{The architecture of the proposed multi-cue emotion recognition approach. Given an input image on an unconstrained scenario, we use an off-the-shelf face detector algorithm \cite{dlib} to get the localization of the face on the image. First, we crop the face and use it as input for the \textit{face encoding stream}, responsible for extracting features from the face. Next, we fill the cropped region with a black rectangle and use this new image as input for the \textit{context encoding stream}. Since the facial crop is occluded, this stream is "forced" to search for features from other image regions during training (i.e., the background context). Finally, we apply a segmentation technique \cite{he2017mask} to remove background noise and persons that are not acting directly on the scene. We use this segmentation mask as input for an off-the-shelf human keypoint extractor \cite{xiao2018simple}. The features extracted from these three streams are fused in an adaptive way allowing the emotion classification.}
	\label{fig:caer}
\end{figure*}

We considered the work of \cite{caer} as a baseline and evolved a new pipeline for emotion recognition. Our approach comprises image preprocessing and face selection strategies, coupled with the usage of self-calibrated convolutions and detection of body keypoints. We display an overview of our architecture in Fig.~\ref{fig:caer}. Based on the excellent results, our main contributions are: \\[-8mm]

\begin{itemize}
\item A body encoding stream that extracts features from body language for emotion recognition, lowering ambiguity and opening a path for in-the-wild emotion recognition
\item A context encoding stream powered by an adaptation of the self-calibrated convolutions \citep{scconv}.  The encoder-like method allows the extraction of more representative features from context information
\item An approach for selecting individual faces in crowded scenes, lowering the impact of the lack of per-person annotations in the evaluated dataset
\item A robust training phase powered by a preprocessing pipeline that keeps important features present in the image and allows for a better generalization during training
\end{itemize}

\section{Background}
\label{sec:background}
\subsection{Body language and emotion}

Nonverbal signals are efficient for communicating affective information between persons due to cues that help recognize emotion. The modern studies in psychology and general human behavior can be traced back to Darwin's book entitled "\textit{The Expression of the Emotions in Man and Animals}", which investigates the role of body movements and facial expressions to communicate emotions \citep{lhommet2014expressing, darwin18721965}. Darwin exemplifies how emotions can act directly on our bodily behavior. Joy, for instance, has a strong tendency to purposeless movements since this feeling causes our blood circulation to accelerate, which stimulates the brain, leading to reactions on the body. On the other hand, pride makes a person exhibit its sense of superiority by holding its head and body erect, making them appear as large as possible, as if, metaphorically, they were puffed up with pride.

The psychology literature tackled the problem of posed emotions against expressed emotions. A few pieces of evidence suggest that these posed expressions are, at least, an approximation to what is felt. However, we may diminish the impact of these problems by using different actors in the experiments that are not aware that body movements are a part of the task, leading them to focus attention on the sentences of the scene \citep{wallbott1986cues, zuckerman1976encoding, Wallbott1998}. \\[-7mm]

\subsection{Emotion recognition}
Most research works focus on recognizing emotion through facial expressions, and they are only able to extract features from facial crops of the target person, showing the limited ability for emotion recognition in the wild \citep{su2021facial, wang2020suppressing, wang2020region, li2018occlusion}. Considering that, to explore these limitations, some works use different visual cues from different input sources \citep{chen2016emotion, kosti2017emotion}. However, there is a lack of solutions focusing on salient parts of the scene that would help the models generalize context information \citep{caer}.

\cite{caer} proposed an approach to deal with this limitation using an attention module that adapts itself and allows a context encoding stream to search for meaningful and salient information on the scene to help recognize and classify emotions. However, although the context encoding stream enables selecting other parts of the scene and even parts of the body, the approach is not trained with specific knowledge on body pose estimation.

Based on the concept that body expression can help the perception of emotions, \cite{randhavane2019identifying} proposes the analysis of gait to classify emotions. Given an RGB video of an individual walking, the authors use 3D pose estimation techniques to create a set of 3D poses, which are spatiotemporally investigated. Although this technique uses information from body language, it requires the user to be walking to extract features such as the swing of the arms and posture, therefore imposing constraints to the user and limiting the range of applicability. A more recent approach \citep{step} improves the previously mentioned technique by 14\% on the accuracy metric but is also limited to the users walking.

\section{MCAER-Net}\label{sec:seamless_approach}

Given an image \(I\), we aim to infer an emotion \(y\) among a set of \(K\) emotion labels by using a convolutional neural network model. The proposed network architecture extracts features of three streams: \textit{face encoding stream}, \textit{context encoding stream} and \textit{body encoding stream}. By combining these features in an \textit{adaptive fusion network}, the proposed method can infer emotion from multiple non-verbal cues. In Fig.~\ref{fig:caer}, we present the proposed architecture, and each module is detailed below.


\subsection{Preprocessing pipeline}. Before our training procedure, we perform a preprocessing step to allow some variability between epochs and enable training while keeping important features available. For the input of the \textit{face encoding stream}, we resize the facial crops to a fixed size of \(96\times96\), following the baseline. For the \textit{context encoding stream}, we pad each image according to the shape of the larger image on the dataset, which is \(400\times712\). We also resize the images by a factor of three to maintain the aspect ratio and use a random crop with a padding of 5 pixels on all sides to augment our training dataset. For the images used in the \textit{body encoding stream}, we follow the pipeline proposed by \cite{xiao2018simple} and resize the image to \(256\times256\).
\label{sub:preprocessing}


\subsection{Multi-cue streams}

\subsubsection{Face encoding stream} 
\begin{figure}
	\centering
	\includegraphics[width=\columnwidth]{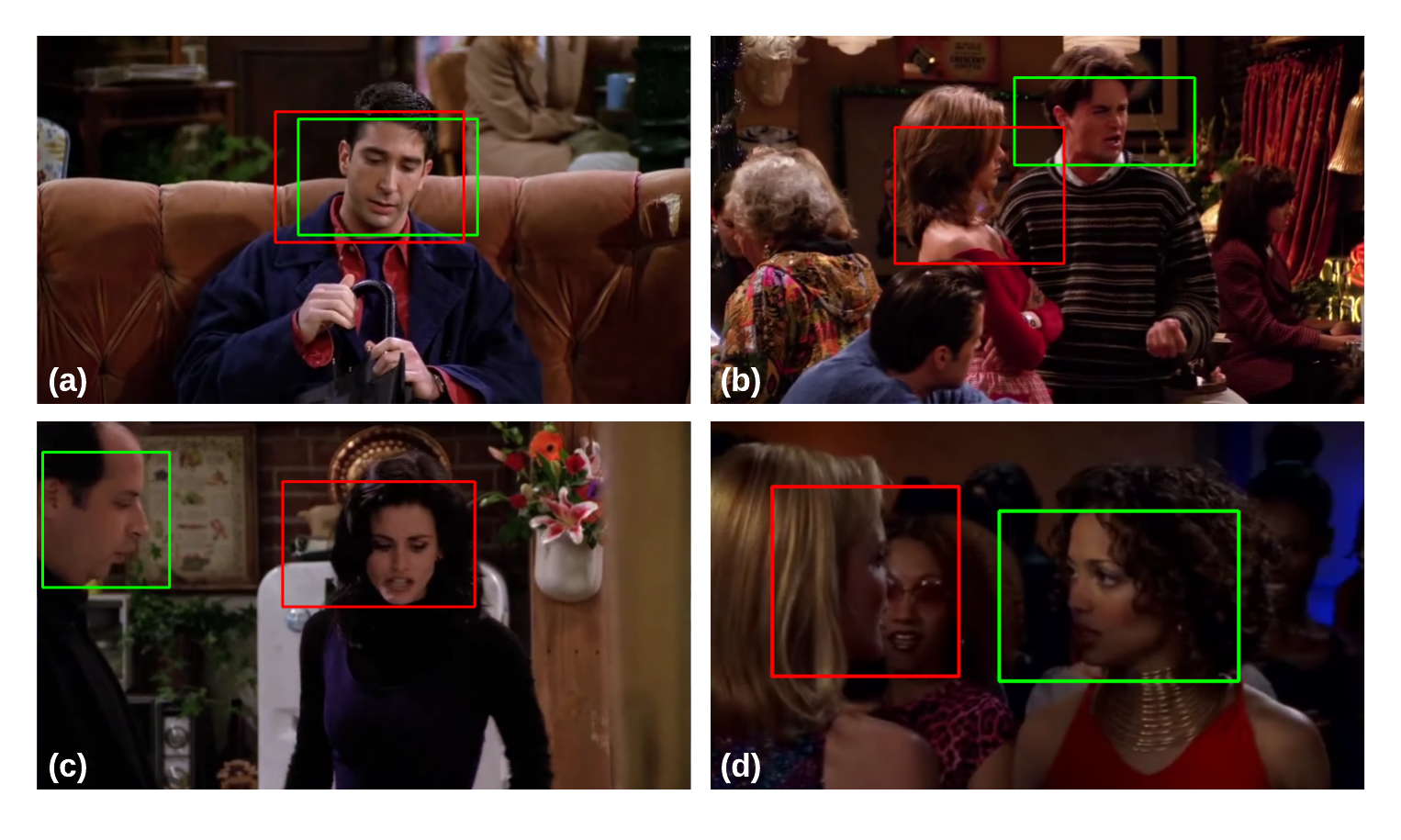}
	\caption{Comparison of different approaches for face crop. In red, the selected faces from the dlib toolkit \citep{dlib}, as proposed by \cite{caer}, and in green, the chosen face using our face selector algorithm. In \textbf{(a)} the only face in the scene is selected. In \textbf{(b)}, the face selector algorithm correctly chooses the face of the principal actor in the TV show scene. The selected face is classified as anger. In \textbf{(c)}, a sample where the algorithm fails to choose the face of the actor. In \textbf{(d)} we show an example of a crowded situation, the face selector algorithm success on selected the principal actress in the scene.}
	\label{fig:caer_face_selector_v2}
\end{figure}

We use an off-the-shelf face detector \citep{dlib} to detect the faces present on the image. In the CAER-S dataset \citep{caer}, the images are extracted from TV shows and contain scenes of interaction between actors; however, there is only a single annotation of emotion on the scene with no remarks of which actor is displaying such emotion. Therefore, if more than one face is present, we use a \textit{face selector algorithm} that selects the principal actor's face based on the estimated distance from the camera (since they are usually in the foreground of the scene) and their position on the image (since they are usually centered on the scene), considering that the given annotation is related to the leading actor. We crop the bounding-box region and use it as input to the \textit{face encoding stream}, as shown in Fig.~\ref{fig:caer}. This module consists of five convolutional layers with 3x3 kernels with sizes 32, 64, 128, 256, and 256, followed by batch normalization (BN) and rectified linear unit (ReLU) activation, and four max-pooling layers with a kernel size of 2. We spatially average the final feature layer using an average-pooling layer. In Fig.~\ref{fig:caer_face_selector_v2}, we show a few examples of the face selector algorithm compared against the original approach.



\subsubsection{Context encoding stream} Extracting emotional information from context is a problematic task due to the high variability of context information. Moreover, many essential details may be hidden in the scene, motivating a robust encoding stream for context. Therefore, proposing approaches to enrich the representations extracted on the \textit{context encoding stream} could improve the results by allowing more representative features to classify emotion. This module consists of four convolutional layers with \(3\times3\) kernels with sizes 32, 64, 128, and 256, followed by BN layers, ReLU activations, and four max-pooling layers with a kernel size of \(2\times2\). Finally, we add an adaptive self-calibrated convolution with kernel size \(3\times3\) and a ReLU activation layer.

\paragraph{Adaptive self-calibrated convolutions} We propose the usage of self-calibrated convolutions \cite{scconv} to allow output features enriched. Those modules provide internal communications of the convolutional layer. It can generate more discriminative representations and improve the overall quality of the extracted features. We replace the last convolutional layer of the convolutional block with a self-calibrated convolution. The adaptive self-calibrated convolution module receives an input with channels size \(C\) and outputs a features map with channels size \(C'\); a restriction when using the original self-calibrated convolutions that we overcame. By altering the last convolution on this block, we allowed encoder networks that vary the output channel size on each convolution based on self-calibrated convolutions.


\paragraph{Attention inference module} An attention inference module is learned in a non-supervised way, enabling the \textit{context encoding stream} to focus on the salient contexts. We use the output features of the \textit{context encoding stream} as input, lowering the number of channels from 236 to 1 and applying a softmax operation to make the sum of attention for each pixel of the input features to be 1. Afterward, the attention is applied to the output feature to make an attention-boosted feature. The intuition beyond the usage of this module is that context may have too much background information, and the attention module helps the \textit{context encoding stream} to select which features are more important, thus boosting its participation in the overall process.

\begin{table*}[t]
\caption{Ablation study of the proposed multi-cue adaptive emotion recognition network.}
\label{table_results_ablation}
\centering
\resizebox{\linewidth}{!}{%
\renewcommand{\arraystretch}{1.3}
\begin{tabular}{c|ccccc|c} 
\toprule
\hline
\multirow{2}{*}{\textbf{Method}} & \multicolumn{6}{c}{\textbf{Evaluated Components}} \\ 
\cline{2-7}
 & \begin{tabular}[c]{@{}c@{}}Self-calibrated convolution on the\\last layer of context encoding stream\end{tabular} & \begin{tabular}[c]{@{}c@{}}Face selector\\algorithm\end{tabular} & \begin{tabular}[c]{@{}c@{}}Preprocessing\\~pipeline\end{tabular} & \begin{tabular}[c]{@{}c@{}}Body encoding\\stream\end{tabular} & \begin{tabular}[c]{@{}c@{}}Body encoding stream\\with segmented body\end{tabular} & Accuracy \\ 
\hline
\multirow{4}{*}{MCAER-Net} & \checkmark & x & x & x & x & 86.70\% \\
 & \checkmark & \checkmark & \checkmark & x & x & 88.10\% \\
 & \checkmark & \checkmark & \checkmark & \checkmark & x & 81.70\% \\
 & \checkmark & \checkmark & \checkmark & x & \checkmark & \textbf{89.30\%} \\
\bottomrule
\hline
\end{tabular}
}
\end{table*}

\subsubsection{Body encoding stream} Following our proposal to investigate the human body pose as a nonverbal communication input, we used an approach proposed by \cite{xiao2018simple} known as SimpleHRNet. The technique is based on the ResNet \citep{resnet} and proposes to add a few deconvolutional layers on the end of the backbone network to estimate heatmaps from feature maps.

\begin{figure}
	\centering
	\includegraphics[width=\columnwidth]{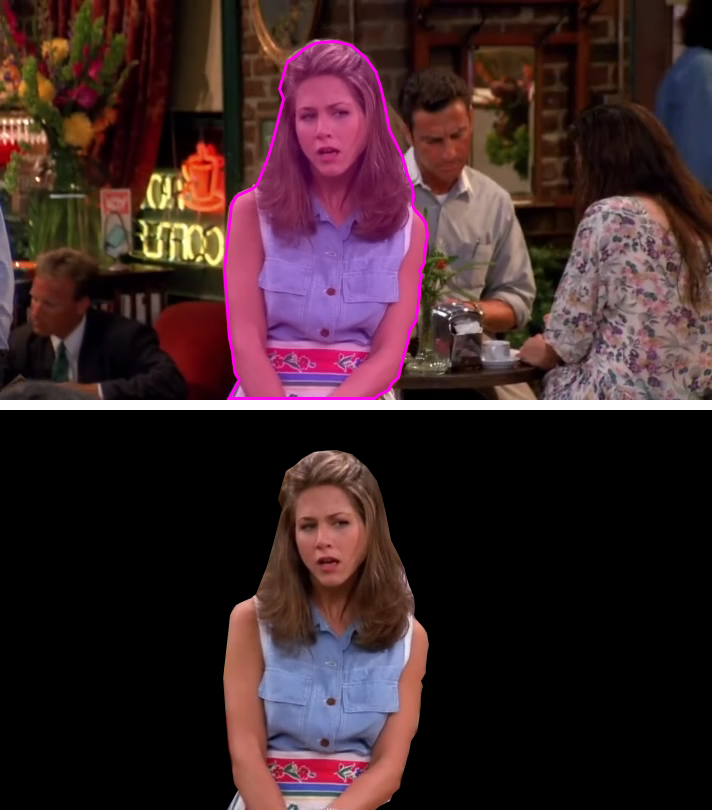}
	\caption{The proposed approach to deal with the cluttered background problem. We regress a mask around the main actor to occlude the background information and force the keypoints detection model to focus on the foreground information.}
	\label{fig:caermask}
\end{figure}

We used the features learned up to the last convolutional layer of the network. However, as described previously, we have more than one interlocutor interacting on a scene in some datasets, and this is a real-world prerequisite. Therefore, we propose to use segmentation techniques to separate the person of interest in the scene and extract only its body pose. Fig.~\ref{fig:caermask} presents an example of how to apply segmentation masks for cluttered backgrounds. We used Mask R-CNN \citep{he2017mask} to create masks of the people present in the scene and calculate the overlap between the selected face from the face selector algorithm and each of the masks. 

This approach prevents that two different persons from being considered on two modules. This way, a refined version of the dataset was preprocessed and saved to eliminate the need to predict each epoch's masks. The \textit{body encoding stream} receives the full, uncropped image, with the removed background, as we show in Fig.~\ref{fig:caer} and Fig.~\ref{fig:caermask}.
\label{para:body}



\begin{figure}[h]
	\centering
	\includegraphics[width=\columnwidth]{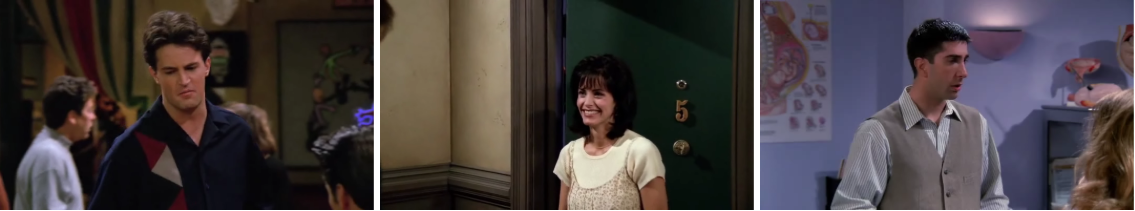}
	\caption{Examples from the CAER-S dataset. From left to right, we present examples of anger, happiness, and surprise.}
	\label{fig:caer_dataset}
\end{figure}

\subsection{Adaptive fusion networks}. To combine face, context, and body pose, we fuse the features extracted from the three modules using an adaptive fusion network with an attention model since the direct concatenation of such different features often fails to provide adequate performance \citep{caer}. According to the contents, the attention weights are determined to self-adapt to yield an optimal solution.

\section{Experiments}\label{sec:experiments}


We implemented MCAER-Net using the PyTorch library \citep{pytorch}. Next, we trained the \textit{face encoding module} and \textit{context encoding module} from scratch with an initial learning rate initialized as \(4\times10^{-3}\) and dropped by a factor of 0.4 every 40 epochs using the RMSProp optimizer. Finally, we trained the model using the cross-entropy loss function on a batch size of 32. 

We use the CAER-S dataset \citep{caer} for experimentation. The dataset is focused on context-aware emotion recognition and is sufficient for the task of multi-cue emotion recognition. The dataset is based on video clips from 79 television shows, and each frame is categorized as one out of seven emotional states, namely: "angry", "disgust", "fear", "happy", "sad", "surprise" and "neutral". Fig.~\ref{fig:caer_dataset} present few examples of the dataset. Each collected clip was annotated manually by three different annotators blindly and independently. The authors used about 70 thousand static images from this dataset to create a subset for the specific task of recognizing emotion on static images. Following the literature, we call this subset CAER-S, which is randomly split into training (70\%), validation (10\%), and testing (20\%) sets.

\section{Results and Discussion}

The quantitative results of each proposed experiment are displayed in Table~\ref{table_results_ablation} and discussed below. All experiments are evaluated using the accuracy metric following literature standards \citep{caer, dhall2016emotiw}.

\begin{figure*}[t!]
	\centering
	\includegraphics[width=\textwidth]{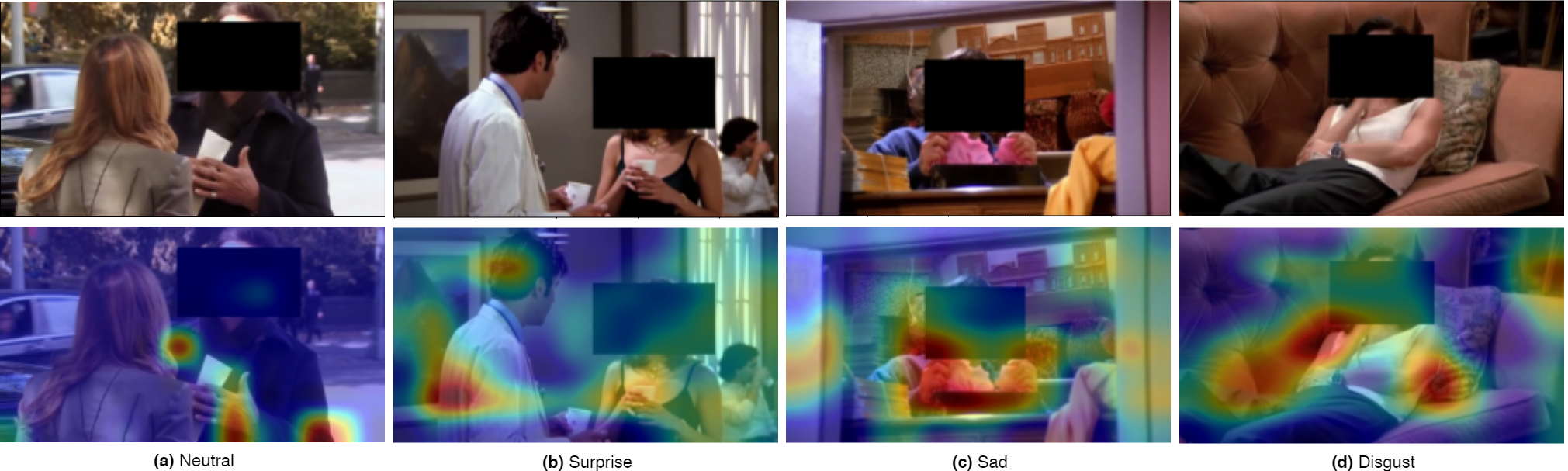}
	\caption{Visualization of the attention module from the context encoding stream. Since this module is responsible for extracting context information, the main person's facial region on the scene is occluded with a black rectangle. On the top row, we see the image used as input for the module, and on the bottom row, the output from Grad-CAM \citep{selvaraju2017grad} with respect to the last convolutional layer of the attention inference module. The red and yellow regions are the most relevant regions that lead to the given prediction. \textbf{(a)} An example of the testing set correctly classified as "neutral." In this example, not much information is extracted from the background, mostly from the main actor's body. \textbf{(b)} An example from the "surprise" subset. In this example, a major part of the main actor's face is considered relevant. \textbf{(c)} In this example from the "sad" subset, the visualization points to the main actor's hands, as if a personal object was torn apart. Finally, in \textbf{(d)} we have an example from the "disgust" subset, pointing to the way that the actor is lying on the couch.}
	\label{fig:caer_gradcam}
\end{figure*}

Our reproduction of the baseline proposed by \cite{caer} yielded an accuracy of 81.5\% in the CAER-S test set. This implementation is our base of comparison against our proposed improvements. Our first proposed experiment is based on the usage of self-calibrated convolutions \citep{scconv} for better extraction of features in the \textit{context encoding stream}. However, swapping the last convolutional layer of both \textit{face encoding stream} and \textit{context encoding stream} prejudiced the overall accuracy. This is due to the size of the face crops, which in a few examples could be relatively small because of the placement of the leading actor on the scene. Thus we proceeded to exchange only the last convolutional layer of the \textit{context encoding stream}, and, as we show in Table~\ref{table_results_ablation}, this experiment led us to a higher accuracy score. We also evaluated exchanging all convolutional layers with the proposed adaptive self-calibrated convolutions. However, we verified a negative impact on the overall accuracy.

Another experiment involved the usage of the \textit{face selector algorithm}. It is not uncommon to have different interlocutors portraying different emotions in a scene with multiple subjects. Therefore, we investigate how a wrongly selected face could impact the overall accuracy. We noticed a slight increase in the accuracy score using this approach. However, it does not significantly contribute to the final accuracy, even selecting the correct face in crowded scenarios. This experiment may point out the robustness of the \textit{context encoding stream} that can leverage emotion from context even with an incorrect face selected. Following the proposed improvements, we evaluated the contribution of the new preprocessing pipeline, introduced in Section~\ref{sub:preprocessing}, to the overall result, which again leads to an increase in inaccuracy.
\begin{table}[b!]
\vspace{-0.5cm}
\centering
\caption{Quantiative evaluation of MCAER-Net in comparison with baseline methods on the CAER-S dataset.}
\label{sota}
\resizebox{\linewidth}{!}{%
\renewcommand{\arraystretch}{1.3}
\begin{tabular}{lr} 
\toprule
Methods                  & \multicolumn{1}{l}{Acc. (\%)}  \\ 
\hline\hline
ImageNet-AlexNet \citep{krizhevsky2012imagenet}        & 47.36                          \\
ImageNet-VGG-Net \citep{simonyan2014very}          & 49.89                          \\
ImageNet-ResNet \citep{he2016deep}         & 57.33                          \\ 
\hline
Fine-tuned AlexNet \citep{krizhevsky2012imagenet}      & 61.73                          \\
Fine-tuned VGGNet \citep{simonyan2014very}       & 64.85                          \\
Fine-tuned ResNet \citep{he2016deep}       & 68.46                          \\ 
\hline
CAER-Net-S  \citep{caer}             & 73.51                         \\ 
CAER-Net-S (our implementation)  \citep{caer}             & 81.50                         \\ 
\hline
MCAER-Net (face+context)           & 86.70                             \\
\textbf{MCAER-Net (face+context+body)} & \textbf{89.30}                            \\
\bottomrule
\end{tabular}
}
\end{table}
Finally, we experiment on the usage of body language as an input to the model. The features learned up to the last convolutional layer are combined on the adaptive fusion module, which learns weights for three inputs. A first experiment focused solely on the usage of the body pose estimation technique as an extra input cue to the model, and, as shown in Table.~\ref{table_results_ablation}, yielded a low accuracy score compared to the previously implemented improvements. However, during further investigation, we noticed that more than one actor is present on the scene in many cases, and the body posture of these actors was also considered when leveraging the body pose. Therefore, as previously explained in Section.~\ref{para:body}, we use Mask R-CNN \citep{he2017mask} to segment the principal actor's body and isolate it from context leading to our best result of 89.3\%. Although the accuracy increment may be considered small compared with the approach using only context and face, the pose encoding stream may be decisive in complex cases where the context does not contain useful information. However, further investigation is needed, specially hyperparameter-wise, to understand if any other constraints contribute negatively to this result.

We compare our results with different baseline approaches in Table~\ref{sota}. The proposed method was 21.49\% better when compared to the highest-scoring approach - CAER-Net-S \cite{caer}. We also performing consistently against traditional deep neural networks approaches for image tasks, such as AlexNet \citep{krizhevsky2012imagenet}, VGG-Net \citep{simonyan2014very}, and ResNet \citep{He_2016}.

\begin{figure}[t!]
	\centering
	\includegraphics[width=\columnwidth]{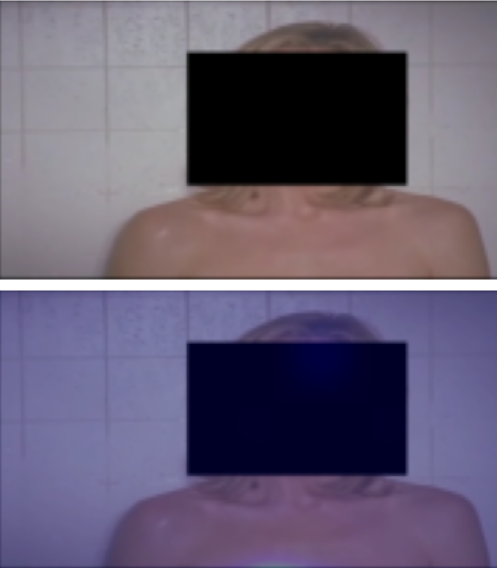}
	\caption{Visualization of an example from the "fear" subset. We see a weak activation map on the image, meaning that the context encoding stream could not extract any information in this case. The proposed technique correctly classified this example, mainly due to the close-up on the face. However, this example would benefit from a \textit{body encoding stream}.}
	\label{fig:caer_negative_pose_b}
\end{figure}

After quantitative evaluation, we proceeded with a visual investigation based on the Grad-CAM technique \citep{selvaraju2017grad}. The Grad-CAM allows visualizations of which regions of the input image are more relevant for predictions by using class-specific gradient information to localize these crucial regions. In Fig.~\ref{fig:caer_gradcam}, we show a few examples of how the context encoding stream acts towards the correct prediction of the emotion. Next, we investigated the specific cases in which the context is not rich in information, such as images containing only white walls in the background. For these scenarios, the technique relies only on the facial features for emotion recognition. It could benefit from more features from the person's posture and body language, as shown in Fig.~\ref{fig:caer_negative_pose_b}. This investigation supports our assumption that relying only on context and face information is not sufficient in some cases, especially on in-the-wild scenarios, in which occlusion of the face is expected, and context may not be representative at all.


\section{Conclusion}\label{sec:conclusion}

In this work, we presented an approach for emotion recognition using multiple cues that represents important non-verbal communication as input. Our proposed model learns to weigh each input adaptively, attributing higher weights to inputs with more descriptive features in each scenario.

Our proposal reaches an accuracy score of 89.30\% on the CAER-S dataset, increasing 21.49\% regarding the current state-of-the-art in the same dataset. Our contributions were supported by research on behavioral psychology and the improvement regarding extraction of features from context and body, among other improvements, open paths towards in-the-wild applications.

\section*{Acknowledgments}
This study was financed in part by the Coordenação de Aperfeiçoamento de Pessoal de Nível Superior - Brasil (CAPES) - Finance Code 001.

\bibliographystyle{model2-names}
\bibliography{refs}

\begin{thebibliography}{40}
\expandafter\ifx\csname natexlab\endcsname\relax\def\natexlab#1{#1}\fi
\providecommand{\url}[1]{\texttt{#1}}
\providecommand{\href}[2]{#2}
\providecommand{\path}[1]{#1}
\providecommand{\DOIprefix}{doi:}
\providecommand{\ArXivprefix}{arXiv:}
\providecommand{\URLprefix}{URL: }
\providecommand{\Pubmedprefix}{pmid:}
\providecommand{\doi}[1]{\href{http://dx.doi.org/#1}{\path{#1}}}
\providecommand{\Pubmed}[1]{\href{pmid:#1}{\path{#1}}}
\providecommand{\bibinfo}[2]{#2}
\ifx\xfnm\relax \def\xfnm[#1]{\unskip,\space#1}\fi
\bibitem[{Bhattacharya et~al.(2020a)Bhattacharya, Mittal, Chandra, Randhavane,
  Bera and Manocha}]{bhattacharya2020step}
\bibinfo{author}{Bhattacharya, U.}, \bibinfo{author}{Mittal, T.},
  \bibinfo{author}{Chandra, R.}, \bibinfo{author}{Randhavane, T.},
  \bibinfo{author}{Bera, A.}, \bibinfo{author}{Manocha, D.},
  \bibinfo{year}{2020}a.
\newblock \bibinfo{title}{Step: Spatial temporal graph convolutional networks
  for emotion perception from gaits}, in: \bibinfo{booktitle}{Proceedings of
  the AAAI Conference on Artificial Intelligence}, pp.
  \bibinfo{pages}{1342--1350}.
\bibitem[{Bhattacharya et~al.(2020b)Bhattacharya, Mittal, Chandra, Randhavane,
  Bera and Manocha}]{step}
\bibinfo{author}{Bhattacharya, U.}, \bibinfo{author}{Mittal, T.},
  \bibinfo{author}{Chandra, R.}, \bibinfo{author}{Randhavane, T.},
  \bibinfo{author}{Bera, A.}, \bibinfo{author}{Manocha, D.},
  \bibinfo{year}{2020}b.
\newblock \bibinfo{title}{Step: Spatial temporal graph convolutional networks
  for emotion perception from gaits}, in: \bibinfo{booktitle}{Proceedings of
  the AAAI Conference on Artificial Intelligence}, pp.
  \bibinfo{pages}{1342--1350}.
\bibitem[{Chen et~al.(2016)Chen, Wu and Jiang}]{chen2016emotion}
\bibinfo{author}{Chen, C.}, \bibinfo{author}{Wu, Z.}, \bibinfo{author}{Jiang,
  Y.G.}, \bibinfo{year}{2016}.
\newblock \bibinfo{title}{Emotion in context: Deep semantic feature fusion for
  video emotion recognition}, in: \bibinfo{booktitle}{Proceedings of the 24th
  ACM international conference on Multimedia}, pp. \bibinfo{pages}{127--131}.
\bibitem[{Crenn et~al.(2016)Crenn, Khan, Meyer and Bouakaz}]{crenn2016body}
\bibinfo{author}{Crenn, A.}, \bibinfo{author}{Khan, R.A.},
  \bibinfo{author}{Meyer, A.}, \bibinfo{author}{Bouakaz, S.},
  \bibinfo{year}{2016}.
\newblock \bibinfo{title}{Body expression recognition from animated 3d
  skeleton}, in: \bibinfo{booktitle}{2016 International Conference on 3D
  Imaging (IC3D)}, \bibinfo{organization}{IEEE}. pp. \bibinfo{pages}{1--7}.
\bibitem[{Darwin(1872)}]{darwin18721965}
\bibinfo{author}{Darwin, C.}, \bibinfo{year}{1872}.
\newblock \bibinfo{title}{1965. the expression of the emotions in man and
  animals}.
\newblock \bibinfo{journal}{London, UK: John Marry} .
\bibitem[{Dhall et~al.(2016)Dhall, Goecke, Joshi, Hoey and
  Gedeon}]{dhall2016emotiw}
\bibinfo{author}{Dhall, A.}, \bibinfo{author}{Goecke, R.},
  \bibinfo{author}{Joshi, J.}, \bibinfo{author}{Hoey, J.},
  \bibinfo{author}{Gedeon, T.}, \bibinfo{year}{2016}.
\newblock \bibinfo{title}{Emotiw 2016: Video and group-level emotion
  recognition challenges}, in: \bibinfo{booktitle}{Proceedings of the 18th ACM
  international conference on multimodal interaction}, pp.
  \bibinfo{pages}{427--432}.
\bibitem[{He et~al.(2017)He, Gkioxari, Doll{\'a}r and Girshick}]{he2017mask}
\bibinfo{author}{He, K.}, \bibinfo{author}{Gkioxari, G.},
  \bibinfo{author}{Doll{\'a}r, P.}, \bibinfo{author}{Girshick, R.},
  \bibinfo{year}{2017}.
\newblock \bibinfo{title}{Mask r-cnn}, in: \bibinfo{booktitle}{Proceedings of
  the IEEE international conference on computer vision}, pp.
  \bibinfo{pages}{2961--2969}.
\bibitem[{He et~al.(2016a)He, Zhang, Ren and Sun}]{resnet}
\bibinfo{author}{He, K.}, \bibinfo{author}{Zhang, X.}, \bibinfo{author}{Ren,
  S.}, \bibinfo{author}{Sun, J.}, \bibinfo{year}{2016}a.
\newblock \bibinfo{title}{Deep residual learning for image recognition}, in:
  \bibinfo{booktitle}{Proceedings of the IEEE conference on computer vision and
  pattern recognition}, pp. \bibinfo{pages}{770--778}.
\bibitem[{He et~al.(2016b)He, Zhang, Ren and Sun}]{he2016deep}
\bibinfo{author}{He, K.}, \bibinfo{author}{Zhang, X.}, \bibinfo{author}{Ren,
  S.}, \bibinfo{author}{Sun, J.}, \bibinfo{year}{2016}b.
\newblock \bibinfo{title}{Deep residual learning for image recognition}, in:
  \bibinfo{booktitle}{Proceedings of the IEEE conference on computer vision and
  pattern recognition}, pp. \bibinfo{pages}{770--778}.
\bibitem[{He et~al.(2016c)He, Zhang, Ren and Sun}]{He_2016}
\bibinfo{author}{He, K.}, \bibinfo{author}{Zhang, X.}, \bibinfo{author}{Ren,
  S.}, \bibinfo{author}{Sun, J.}, \bibinfo{year}{2016}c.
\newblock \bibinfo{title}{Identity mappings in deep residual networks}.
\newblock \bibinfo{journal}{European Conference on Computer Vision} .
\bibitem[{King(2009)}]{dlib}
\bibinfo{author}{King, D.E.}, \bibinfo{year}{2009}.
\newblock \bibinfo{title}{Dlib-ml: A machine learning toolkit}.
\newblock \bibinfo{journal}{The Journal of Machine Learning Research}
  \bibinfo{volume}{10}, \bibinfo{pages}{1755--1758}.
\bibitem[{{Kleinsmith} and {Bianchi-Berthouze}(2013)}]{andreaaffective}
\bibinfo{author}{{Kleinsmith}, A.}, \bibinfo{author}{{Bianchi-Berthouze}, N.},
  \bibinfo{year}{2013}.
\newblock \bibinfo{title}{Affective body expression perception and recognition:
  A survey}.
\newblock \bibinfo{journal}{IEEE Transactions on Affective Computing}
  \bibinfo{volume}{4}, \bibinfo{pages}{15--33}.
\newblock \DOIprefix\doi{10.1109/T-AFFC.2012.16}.
\bibitem[{Kosti et~al.(2017)Kosti, Alvarez, Recasens and
  Lapedriza}]{kosti2017emotion}
\bibinfo{author}{Kosti, R.}, \bibinfo{author}{Alvarez, J.M.},
  \bibinfo{author}{Recasens, A.}, \bibinfo{author}{Lapedriza, A.},
  \bibinfo{year}{2017}.
\newblock \bibinfo{title}{Emotion recognition in context}, in:
  \bibinfo{booktitle}{Proceedings of the IEEE conference on computer vision and
  pattern recognition}, pp. \bibinfo{pages}{1667--1675}.
\bibitem[{Kosti et~al.(2019)Kosti, Alvarez, Recasens and
  Lapedriza}]{kosti2019context}
\bibinfo{author}{Kosti, R.}, \bibinfo{author}{Alvarez, J.M.},
  \bibinfo{author}{Recasens, A.}, \bibinfo{author}{Lapedriza, A.},
  \bibinfo{year}{2019}.
\newblock \bibinfo{title}{Context based emotion recognition using emotic
  dataset}.
\newblock \bibinfo{journal}{IEEE transactions on pattern analysis and machine
  intelligence} \bibinfo{volume}{42}, \bibinfo{pages}{2755--2766}.
\bibitem[{Krizhevsky et~al.(2012)Krizhevsky, Sutskever and
  Hinton}]{krizhevsky2012imagenet}
\bibinfo{author}{Krizhevsky, A.}, \bibinfo{author}{Sutskever, I.},
  \bibinfo{author}{Hinton, G.E.}, \bibinfo{year}{2012}.
\newblock \bibinfo{title}{Imagenet classification with deep convolutional
  neural networks}.
\newblock \bibinfo{journal}{Advances in neural information processing systems}
  \bibinfo{volume}{25}, \bibinfo{pages}{1097--1105}.
\bibitem[{Lee et~al.(2019)Lee, Kim, Kim, Park and Sohn}]{caer}
\bibinfo{author}{Lee, J.}, \bibinfo{author}{Kim, S.}, \bibinfo{author}{Kim,
  S.}, \bibinfo{author}{Park, J.}, \bibinfo{author}{Sohn, K.},
  \bibinfo{year}{2019}.
\newblock \bibinfo{title}{Context-aware emotion recognition networks}, in:
  \bibinfo{booktitle}{Proceedings of the IEEE/CVF International Conference on
  Computer Vision}, pp. \bibinfo{pages}{10143--10152}.
\bibitem[{Lhommet and Marsella(2014)}]{lhommet2014expressing}
\bibinfo{author}{Lhommet, M.}, \bibinfo{author}{Marsella, S.C.},
  \bibinfo{year}{2014}.
\newblock \bibinfo{title}{Expressing emotion through posture}.
\newblock \bibinfo{journal}{The Oxford handbook of affective computing}
  \bibinfo{volume}{273}.
\bibitem[{Li et~al.(2018)Li, Zeng, Shan and Chen}]{li2018occlusion}
\bibinfo{author}{Li, Y.}, \bibinfo{author}{Zeng, J.}, \bibinfo{author}{Shan,
  S.}, \bibinfo{author}{Chen, X.}, \bibinfo{year}{2018}.
\newblock \bibinfo{title}{Occlusion aware facial expression recognition using
  cnn with attention mechanism}.
\newblock \bibinfo{journal}{IEEE Transactions on Image Processing}
  \bibinfo{volume}{28}, \bibinfo{pages}{2439--2450}.
\bibitem[{Liu et~al.(2020)Liu, Hou, Cheng, Wang and Feng}]{scconv}
\bibinfo{author}{Liu, J.J.}, \bibinfo{author}{Hou, Q.}, \bibinfo{author}{Cheng,
  M.M.}, \bibinfo{author}{Wang, C.}, \bibinfo{author}{Feng, J.},
  \bibinfo{year}{2020}.
\newblock \bibinfo{title}{Improving convolutional networks with self-calibrated
  convolutions}, in: \bibinfo{booktitle}{Proceedings of the IEEE/CVF Conference
  on Computer Vision and Pattern Recognition}, pp.
  \bibinfo{pages}{10096--10105}.
\bibitem[{{Lucey} et~al.(2010){Lucey}, {Cohn}, {Kanade}, {Saragih}, {Ambadar}
  and {Matthews}}]{ck}
\bibinfo{author}{{Lucey}, P.}, \bibinfo{author}{{Cohn}, J.F.},
  \bibinfo{author}{{Kanade}, T.}, \bibinfo{author}{{Saragih}, J.},
  \bibinfo{author}{{Ambadar}, Z.}, \bibinfo{author}{{Matthews}, I.},
  \bibinfo{year}{2010}.
\newblock \bibinfo{title}{The extended cohn-kanade dataset (ck+): A complete
  dataset for action unit and emotion-specified expression}, in:
  \bibinfo{booktitle}{2010 IEEE Computer Society Conference on Computer Vision
  and Pattern Recognition - Workshops}, pp. \bibinfo{pages}{94--101}.
\newblock \DOIprefix\doi{10.1109/CVPRW.2010.5543262}.
\bibitem[{Mar{\'\i}n-Morales et~al.(2018)Mar{\'\i}n-Morales, Higuera-Trujillo,
  Greco, Guixeres, Llinares, Scilingo, Alca{\~n}iz and Valenza}]{ref1_vr}
\bibinfo{author}{Mar{\'\i}n-Morales, J.}, \bibinfo{author}{Higuera-Trujillo,
  J.L.}, \bibinfo{author}{Greco, A.}, \bibinfo{author}{Guixeres, J.},
  \bibinfo{author}{Llinares, C.}, \bibinfo{author}{Scilingo, E.P.},
  \bibinfo{author}{Alca{\~n}iz, M.}, \bibinfo{author}{Valenza, G.},
  \bibinfo{year}{2018}.
\newblock \bibinfo{title}{Affective computing in virtual reality: emotion
  recognition from brain and heartbeat dynamics using wearable sensors}.
\newblock \bibinfo{journal}{Scientific reports} \bibinfo{volume}{8},
  \bibinfo{pages}{1--15}.
\bibitem[{Paszke et~al.(2017)Paszke, Gross, Chintala, Chanan, Yang, DeVito,
  Lin, Desmaison, Antiga and Lerer}]{pytorch}
\bibinfo{author}{Paszke, A.}, \bibinfo{author}{Gross, S.},
  \bibinfo{author}{Chintala, S.}, \bibinfo{author}{Chanan, G.},
  \bibinfo{author}{Yang, E.}, \bibinfo{author}{DeVito, Z.},
  \bibinfo{author}{Lin, Z.}, \bibinfo{author}{Desmaison, A.},
  \bibinfo{author}{Antiga, L.}, \bibinfo{author}{Lerer, A.},
  \bibinfo{year}{2017}.
\newblock \bibinfo{title}{Automatic differentiation in pytorch} .
\bibitem[{Patel(2014)}]{patel2014body}
\bibinfo{author}{Patel, D.S.}, \bibinfo{year}{2014}.
\newblock \bibinfo{title}{Body language: An effective communication tool.}
\newblock \bibinfo{journal}{IUP Journal of English Studies}
  \bibinfo{volume}{9}.
\bibitem[{Randhavane et~al.(2019a)Randhavane, Bhattacharya, Kapsaskis, Gray,
  Bera and Manocha}]{randhavane2019identifying}
\bibinfo{author}{Randhavane, T.}, \bibinfo{author}{Bhattacharya, U.},
  \bibinfo{author}{Kapsaskis, K.}, \bibinfo{author}{Gray, K.},
  \bibinfo{author}{Bera, A.}, \bibinfo{author}{Manocha, D.},
  \bibinfo{year}{2019}a.
\newblock \bibinfo{title}{Identifying emotions from walking using affective and
  deep features}.
\newblock \bibinfo{journal}{arXiv preprint arXiv:1906.11884} .
\bibitem[{Randhavane et~al.(2019b)Randhavane, Bhattacharya, Kapsaskis, Gray,
  Bera and Manocha}]{randhavane2019liar}
\bibinfo{author}{Randhavane, T.}, \bibinfo{author}{Bhattacharya, U.},
  \bibinfo{author}{Kapsaskis, K.}, \bibinfo{author}{Gray, K.},
  \bibinfo{author}{Bera, A.}, \bibinfo{author}{Manocha, D.},
  \bibinfo{year}{2019}b.
\newblock \bibinfo{title}{The liar's walk: Detecting deception with gait and
  gesture}.
\newblock \bibinfo{journal}{arXiv preprint arXiv:1912.06874} .
\bibitem[{{Rouast} et~al.(2018){Rouast}, {Adam} and {Chiong}}]{rouast2019deep}
\bibinfo{author}{{Rouast}, P.V.}, \bibinfo{author}{{Adam}, M.},
  \bibinfo{author}{{Chiong}, R.}, \bibinfo{year}{2018}.
\newblock \bibinfo{title}{Deep learning for human affect recognition: Insights
  and new developments}.
\newblock \bibinfo{journal}{IEEE Transactions on Affective Computing} ,
  \bibinfo{pages}{1--1}\DOIprefix\doi{10.1109/TAFFC.2018.2890471}.
\bibitem[{Selvaraju et~al.(2017)Selvaraju, Cogswell, Das, Vedantam, Parikh and
  Batra}]{selvaraju2017grad}
\bibinfo{author}{Selvaraju, R.R.}, \bibinfo{author}{Cogswell, M.},
  \bibinfo{author}{Das, A.}, \bibinfo{author}{Vedantam, R.},
  \bibinfo{author}{Parikh, D.}, \bibinfo{author}{Batra, D.},
  \bibinfo{year}{2017}.
\newblock \bibinfo{title}{Grad-cam: Visual explanations from deep networks via
  gradient-based localization}, in: \bibinfo{booktitle}{Proceedings of the IEEE
  international conference on computer vision}, pp. \bibinfo{pages}{618--626}.
\bibitem[{Setiono et~al.(2021)Setiono, Saputra, Putra, Moniaga and
  Chowanda}]{ref1_enter}
\bibinfo{author}{Setiono, D.}, \bibinfo{author}{Saputra, D.},
  \bibinfo{author}{Putra, K.}, \bibinfo{author}{Moniaga, J.V.},
  \bibinfo{author}{Chowanda, A.}, \bibinfo{year}{2021}.
\newblock \bibinfo{title}{Enhancing player experience in game with affective
  computing}.
\newblock \bibinfo{journal}{Procedia Computer Science} \bibinfo{volume}{179},
  \bibinfo{pages}{781--788}.
\newblock \URLprefix
  \url{https://www.sciencedirect.com/science/article/pii/S1877050921000843},
  \DOIprefix\doi{https://doi.org/10.1016/j.procs.2021.01.066}.
  \bibinfo{note}{5th International Conference on Computer Science and
  Computational Intelligence 2020}.
\bibitem[{Simonyan and Zisserman(2014)}]{simonyan2014very}
\bibinfo{author}{Simonyan, K.}, \bibinfo{author}{Zisserman, A.},
  \bibinfo{year}{2014}.
\newblock \bibinfo{title}{Very deep convolutional networks for large-scale
  image recognition}.
\newblock \bibinfo{journal}{arXiv preprint arXiv:1409.1556} .
\bibitem[{Su et~al.(2021)Su, Zhang, Su and Yu}]{su2021facial}
\bibinfo{author}{Su, W.}, \bibinfo{author}{Zhang, H.}, \bibinfo{author}{Su,
  Y.}, \bibinfo{author}{Yu, J.}, \bibinfo{year}{2021}.
\newblock \bibinfo{title}{Facial expression recognition with confidence guided
  refined horizontal pyramid network}.
\newblock \bibinfo{journal}{IEEE Access} \bibinfo{volume}{9},
  \bibinfo{pages}{50321--50331}.
\bibitem[{Valli(2008)}]{valli2008}
\bibinfo{author}{Valli, A.}, \bibinfo{year}{2008}.
\newblock \bibinfo{title}{{Alessandro Valli - Notes on Natural Interaction}}.
\bibitem[{Wallbott(1998)}]{Wallbott1998}
\bibinfo{author}{Wallbott, H.G.}, \bibinfo{year}{1998}.
\newblock \bibinfo{title}{{Bodily expression of emotion}}.
\newblock \bibinfo{journal}{European Journal of Social Psychology}
  \bibinfo{volume}{28}, \bibinfo{pages}{879--896}.
\newblock
  \DOIprefix\doi{10.1002/(SICI)1099-0992(1998110)28:6<879::AID-EJSP901>3.0.CO;2-W}.
\bibitem[{Wallbott and Scherer(1986)}]{wallbott1986cues}
\bibinfo{author}{Wallbott, H.G.}, \bibinfo{author}{Scherer, K.R.},
  \bibinfo{year}{1986}.
\newblock \bibinfo{title}{Cues and channels in emotion recognition.}
\newblock \bibinfo{journal}{Journal of personality and social psychology}
  \bibinfo{volume}{51}, \bibinfo{pages}{690}.
\bibitem[{Wang et~al.(2020a)Wang, Peng, Yang, Lu and
  Qiao}]{wang2020suppressing}
\bibinfo{author}{Wang, K.}, \bibinfo{author}{Peng, X.}, \bibinfo{author}{Yang,
  J.}, \bibinfo{author}{Lu, S.}, \bibinfo{author}{Qiao, Y.},
  \bibinfo{year}{2020}a.
\newblock \bibinfo{title}{Suppressing uncertainties for large-scale facial
  expression recognition}, in: \bibinfo{booktitle}{Proceedings of the IEEE/CVF
  Conference on Computer Vision and Pattern Recognition}, pp.
  \bibinfo{pages}{6897--6906}.
\bibitem[{Wang et~al.(2020b)Wang, Peng, Yang, Meng and Qiao}]{wang2020region}
\bibinfo{author}{Wang, K.}, \bibinfo{author}{Peng, X.}, \bibinfo{author}{Yang,
  J.}, \bibinfo{author}{Meng, D.}, \bibinfo{author}{Qiao, Y.},
  \bibinfo{year}{2020}b.
\newblock \bibinfo{title}{Region attention networks for pose and occlusion
  robust facial expression recognition}.
\newblock \bibinfo{journal}{IEEE Transactions on Image Processing}
  \bibinfo{volume}{29}, \bibinfo{pages}{4057--4069}.
\bibitem[{Xiao et~al.(2018)Xiao, Wu and Wei}]{xiao2018simple}
\bibinfo{author}{Xiao, B.}, \bibinfo{author}{Wu, H.}, \bibinfo{author}{Wei,
  Y.}, \bibinfo{year}{2018}.
\newblock \bibinfo{title}{Simple baselines for human pose estimation and
  tracking}, in: \bibinfo{booktitle}{Proceedings of the European conference on
  computer vision (ECCV)}, pp. \bibinfo{pages}{466--481}.
\bibitem[{Yadegaridehkordi et~al.(2019)Yadegaridehkordi, Noor, Ayub, Affal and
  Hussin}]{ref1_edu}
\bibinfo{author}{Yadegaridehkordi, E.}, \bibinfo{author}{Noor, N.F.B.M.},
  \bibinfo{author}{Ayub, M.N.B.}, \bibinfo{author}{Affal, H.B.},
  \bibinfo{author}{Hussin, N.B.}, \bibinfo{year}{2019}.
\newblock \bibinfo{title}{Affective computing in education: A systematic review
  and future research}.
\newblock \bibinfo{journal}{Computers \& Education} \bibinfo{volume}{142},
  \bibinfo{pages}{103649}.
\newblock \URLprefix
  \url{https://www.sciencedirect.com/science/article/pii/S0360131519302027},
  \DOIprefix\doi{https://doi.org/10.1016/j.compedu.2019.103649}.
\bibitem[{Yannakakis(2018)}]{ref1_health}
\bibinfo{author}{Yannakakis, G.N.}, \bibinfo{year}{2018}.
\newblock \bibinfo{title}{Enhancing health care via affective computing} .
\bibitem[{{Zucco} et~al.(2017){Zucco}, {Calabrese} and
  {Cannataro}}]{ref1_mentalhealth}
\bibinfo{author}{{Zucco}, C.}, \bibinfo{author}{{Calabrese}, B.},
  \bibinfo{author}{{Cannataro}, M.}, \bibinfo{year}{2017}.
\newblock \bibinfo{title}{Sentiment analysis and affective computing for
  depression monitoring}, in: \bibinfo{booktitle}{2017 IEEE International
  Conference on Bioinformatics and Biomedicine (BIBM)}, pp.
  \bibinfo{pages}{1988--1995}.
\newblock \DOIprefix\doi{10.1109/BIBM.2017.8217966}.
\bibitem[{Zuckerman et~al.(1976)Zuckerman, Hall, DeFrank and
  Rosenthal}]{zuckerman1976encoding}
\bibinfo{author}{Zuckerman, M.}, \bibinfo{author}{Hall, J.A.},
  \bibinfo{author}{DeFrank, R.S.}, \bibinfo{author}{Rosenthal, R.},
  \bibinfo{year}{1976}.
\newblock \bibinfo{title}{Encoding and decoding of spontaneous and posed facial
  expressions.}
\newblock \bibinfo{journal}{Journal of Personality and Social Psychology}
  \bibinfo{volume}{34}, \bibinfo{pages}{966}.

\end{thebibliography}
\end{document}